\documentclass{bmvc2k}

\usepackage{booktabs} 
\usepackage{amsmath}
\usepackage{amssymb} 

\newcommand{\rev}[1]{{#1}}

\title{Learning multiplane images from single views with self-supervision}

\addauthor{Gustavo Sutter P. Carvalho$^\star$}{}{1}
\addauthor{Diogo C. Luvizon$^\star$}{}{1}
\addauthor{Antonio Joia Neto}{}{1}
\addauthor{Andr\'{e} G. C. Pacheco}{}{1}
\addauthor{Ot\'{a}vio A. B. Penatti}{}{1}

\addinstitution{
  AI R\&D Lab\\
  Samsung R\&D Institute Brazil\\
  Campinas, SP, 13097-160, Brazil
}

\runninghead{Carvalho et al.}{Learning MPI from single views with self-supervision}

\begin{document}

\maketitle

\begin{abstract}
Generating \rev{static} novel views from an already captured image is a hard task
in computer vision and graphics, in particular when the single input image
has dynamic parts such as persons or moving objects.
In this paper, we tackle this problem by proposing a new framework, called CycleMPI,
that is capable of learning a multiplane image representation from single
images through a cyclic training strategy for self-supervision. Our framework does not require
stereo data for training, therefore it can be trained with massive visual data
from the Internet, resulting in a better generalization capability even
for very challenging cases. Although our method does not require stereo data for
supervision, it reaches results on stereo datasets comparable to the state
of the art in a zero-shot scenario.
We evaluated our method on RealEstate10K and \rev{Mannequin} Challenge
datasets for view synthesis and presented qualitative results on Places II dataset.
\end{abstract}

\section{Introduction}

Predicting novel views from a single image is a key problem
in computer vision and graphics, with potential applications
related to image editing, photo animation, robotics, and virtual reality.
This challenging task requires estimating the scene geometry
while filling in occluded regions in the source image.
For this reason, most of the current solutions require stereo
data during inference
\cite{Srinivasan_CVPR_2019, lsiTulsiani18, mildenhall2020nerf},
which eases the process of estimating the depth of the scene and
helps inpainting occluded regions by gathering relevant
visual information from different points of view.
However, in many applications such as animating an old photo,
multiple views are not available during inference.

Even during the training stage, multiple views of the same scene
are not easy to obtain. Despite the fact that state-of-the-art
neural network models require a large number of images in varied
conditions for training, capturing a massive amount of stereo data
with calibrated and synchronized cameras is an onerous process.
As an alternative, structure-from-motion (SfM) and
Multi-View Stereo (MVS) approaches such as
COLMAP~\cite{schonberger_cvpr_2016} have been used to
estimate the camera parameters from videos~\cite{zhou2018stereo}.
However, since MVS depends on correspondence matching at different
frames, MVS-based approaches might not be suitable for dynamic scenes
with moving people and objects. Considering this limitation,
one of the goals of our method is to learn to generate \rev{static} novel views
from challenging scenes, \rev{including snapshots of scenes with dynamic content or moving people}, without explicitly depending on stereo data for it.

Although a few methods have been proposed to tackle novel view
synthesis from a single image, they are based on
point-cloud~\cite{Niklaus_TOG_2019} or shape
representations~\cite{Shih3DP20}, which require rendering techniques
that are not easily integrated into current deep learning frameworks,
apart from limiting possible applications for the same reason, since
some target devices have limited 3D rendering capabilities.
As an alternative, multiplane images (MPI)~\cite{szeliski1999stereo}
have been proposed to allow rendering of apparent 3D effects from
a set of 2D images.

In this paper, we present CycleMPI, a new framework for learning an MPI
representation in a self-supervised manner, which does not
require multi-view stereo data for training. Instead of relying on multiple
views of the same scene for supervision, we propose to learn an MPI
with a cycle consistency supervision, as illustrated in Fig.~\ref{fig:proposed-method},
which depends only on single images and their correspondent (estimated) depth maps.
Our method can be trained with massive data collected from the Internet, resulting
in a robust solution even for zero-shot evaluation, as \rev{demonstrated} by our results on challenging benchmarks.

The main contributions of our work are summarized as follows. We propose a new MPI learning framework based on cycle consistency loss. Through an extensive experimental evaluation, we demonstrated that our method can be trained with single view images and estimated depth maps. Our approach achieved results close to the state of the art on RealEstate10K in a zero-shot scenario, even when comparing to methods trained on this dataset or using stereo depth predictions during inference.
In the remaining of this paper, we
present a review of the literature in Section~\ref{sec:relatedwork}.
In Section~\ref{sec:method}, we present our method. The experiments and our
conclusions are presented in Sections~\ref{sec:experiments} and~\ref{sec:conclusion}, respectively.

\section{Related work}
\label{sec:relatedwork}

In this section, we go through some of the most relevant work related to our method.

\noindent
\textbf{Single-image view synthesis}.
The goal of single-image view synthesis is to generate images of a scene from novel viewpoints using only one image as input. It states a harder problem than interpolating between two or more views of the same scene, given that occluded information that might be revealed on the multi-image setting is not available on single-image approaches. Tucker and Snavely~\cite{Tucker_2020_CVPR} proposed to learn an MPI representation from a stereo pair of images and an additional depth supervision. Despite \rev{achieving} good results on static scenes, their method still requires multi-view stereo data for training, which hinders the generalization capability.
Huang et al.~\cite{Huang_ECCV2020} proposed a view synthesis method based on generative networks. In their approach, a scene is artificially generated from segmentation masks. Differently, in our work, we target to generate an MPI from an existing photo.
Other more \textit{ad hoc} methods generate an intermediate representation based on a predefined sequence of steps, involving depth estimation, image inpainting, and additional filtering stages~\cite{Shih3DP20, luvizon2021adaptive}. This strategy usually results in better generalization, but has the drawback of requiring an elevated computational cost during inference and complex designing requirements.
A common characteristic of previous methods, as in~\cite{Tucker_2020_CVPR, Shih3DP20}, is the requirement for precise depth maps during training, which are generally estimated by structure-from-motion.

\noindent
\textbf{Monocular depth estimation}.
Single-image view synthesis is highly dependent on the depth of the scene, as previously noted in~\cite{Tucker_2020_CVPR}.
In recent years, many works tackled monocular depth estimation with leaning-based methods, \rev{in particular} deep neural networks. The use of standard convolutional networks with a scale-invariant loss function was first introduced by Eigen et al.~\cite{eigen2014depth}, followed by the use of more complex methods and architectures that further improved the results~\cite{chen2016depthwild, MegaDepthLi18, Alhashim2018}. To deal with poor generalization, Ranftl et al.~\cite{Ranftl2020} combined multiple datasets using a flexible loss function, resulting in impressive zero-shot performance. \rev{Recently}, methods based on the Vision Transfomer (ViT) architecture~\cite{dosovitskiy2020image} have achieved state-of-the-art results for the task~\cite{Ranftl2021}~\cite{bhat2020adabins}.
\rev{Contemporary methods use monocular depth estimation to improve robustness and generalization of stereo depth estimation~\cite{watson-2020-stereo-from-mono} and optical flow~\cite{aleotti2021learning}. In a similar fashion, our work explore monocular depth estimation to improve the generalization of novel view synthesis using MPI representation.}

\noindent
\textbf{Cycle consistency loss}.
The use of cycle consistency loss, also referred to as reconstruction loss in some contexts, has recently gained relevance as \rev{an} objective function in unsupervised and self-supervised problems.
In general, this framework is used to learn transformations over the original data without having necessarily a set of transformed samples for supervision. In order to obtain a supervision signal, the inverse transformation is applied to the intermediate representations, allowing for the reconstruction of the input samples.
In this direction, cycle consistency has been used for image generation with conditional GANs, where the learned transformation is the mapping from one object category to another~\cite{zhu2017unpaired, kim2017learning} or even in a many-to-many fashion~\cite{choi2018stargan}.
Hani et al~\cite{hani2020object} use a consistency loss on the features of the learned transformation to perform novel-view synthesis of 3D shapes without a ground-truth 3D model. Their method, however, diverges from ours by using a neural scene representation and needing two images of \rev{an} object during training.
In our method, we explore the idea of cycle consistency loss to allow a self-supervised training strategy that enables our network to learn inpainting information from single-view images.



\section{Method}
\label{sec:method}

The goal of our method is to synthesize novel views from a single image, through an intermediate MPI representation, without requiring stereo data during training. For this, we propose a neural network that predicts an MPI from a single view. Our neural network is supervised by two main components: depth estimation loss and visual consistency loss. The depth estimation loss can be supervised with predictions (pseudo ground-truth) from a teacher model. This prevents our method from requiring data with precise and dense depth maps, which is hard to obtain on large scale and in varied conditions. The visual consistency is enforced by image losses, which also include inverse projection and cyclic losses. In this section, we briefly review the MPI representation and present our framework in \rev{detail}.

\subsection{Background on multiplane images}
\label{sec:background-mpi}

A \textit{multiplane image} (MPI)~\cite{szeliski1999stereo, zhou2018stereo} can be defined as a set of front-parallel planes, where each plane is an RGBA image, i.e., it encodes color and transparency, and is placed at a fixed depth w.r.t. the camera. Different from other representations, like LDI~\cite{Shade_LDI_1998} or point-cloud~\cite{Niklaus_TOG_2019}, MPI is capable of encoding complex scenes, including transparency and soft edges, can be easily rendered in a fully differentiable way, and benefits real-time applications.

\begin{figure}[!ht]
\centering
    \includegraphics[width=\textwidth]{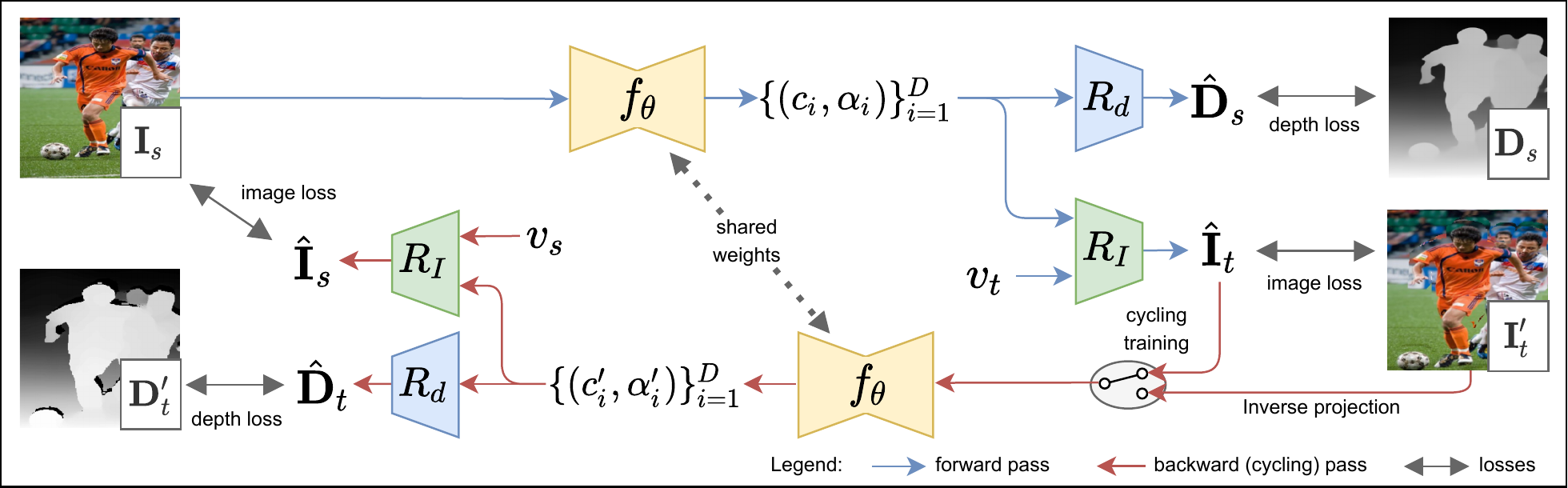}
\caption{Proposed CycleMPI: a source image $\mathbf{I}_s$ is fed into a neural network $f_{\theta}$, which produces an MPI representation w.r.t. the \textit{source viewpoint} $v_s$. This MPI is then rendered in the target viewpoint $v_t$, resulting in $\hat{\mathbf{I}}_t$, which can be fed back into the network. The new MPI produced w.r.t. to the \textit{target viewpoint} is then rendered in the source viewpoint, closing the loop for cyclic supervision. Depth maps are used as additional supervisions.}
\label{fig:proposed-method}
\end{figure}

\noindent
\textbf{MPI representation}.
Given an input image $\mathbf{I}$ of shape $H\times{}W\times{}3$ and a neural network $f_{\theta}$, a multiplane image is obtained by:
\begin{equation}
    \{(c_i, \alpha_i)\}_{i=1}^D=f_{\theta}(\mathbf{I}),
    \label{eq:mpi-gen}
\end{equation}
where $c_i$ and $\alpha_i$ represent the RGB color and alpha values of the $i^{\text{th}}$ image plane, respectively, and $D$ is the total number of planes, assuming $i=1$ the farthest plane. Since directly predicting the RGBA values for each image plane would be highly over-parameterized, it is more common to predict a background image and the alpha values of each image plane, as in~\cite{zhou2018stereo}. In such a manner, the colors of each plane are defined by a combination of the source image $\mathbf{I}$ and the predicted background image $\hat{\mathbf{I}}_b$:
\begin{equation}
    c_i=w_i\odot{}\mathbf{I} + (1-w_i)\odot{}\hat{\mathbf{I}}_b,
\end{equation}
where $\odot{}$ is the Hadamard product. The blending weights $w_i$ are computed from $\alpha_i$~\cite{Tucker_2020_CVPR}:
\begin{equation}
    w_i=\alpha{}_i\prod_{j=i+1}^{D}(1-\alpha{}_j),
    \label{eq:blendingweights}
\end{equation}
which is essentially the over-composite operation in the alphas. Therefore, the actual output of the neural network can be simply the background image $\hat{\mathbf{I}}_b$ and the alpha values $\{\alpha_i\}_{i=1}^D$.

\noindent
\textbf{Warping}.
The MPI representation can be rendered at different points of view through warping and compositing processes.
Warping is implemented as a standard inverse homography~\cite{Hartley2004}, where each image plane is projected from the source to a target viewpoint, considering a given intrinsic and extrinsic camera parameters and the depth $d_i$ of each image plane. More specifically, if we consider the warping operation $\mathcal{W}_{v_s,v_t}$ from the source viewpoint $v_s$ to the target viewpoint $v_t$, the projected color and alpha planes are given by:
\begin{equation}
    c'_i = \mathcal{W}_{v_s,v_t}(d_i, c_i),\quad%
    \alpha'_i = \mathcal{W}_{v_s,v_t}(d_i, \alpha_i).
\end{equation}
\rev{Since we consider images from unknown cameras and MPI layers at fixed depths, we can assume standard intrinsics based on the image size and virtual viewpoints with normalized camera pose.}
For more details about the warping process, please refer to \cite{zhou2018stereo} and \cite{Tucker_2020_CVPR}.

\noindent
\textbf{Compositing}.
After warping, an MPI can be rendered into a target viewpoint $v_t$ by the over-composite operation~\cite{Porter_OverComposite_ACM_1984}. As noted in~\cite{Tucker_2020_CVPR}, the over-compositing operation can be performed on both image and depth domains. In the latter, a depth map can be directly regressed from the alpha channels, which is particularly useful \rev{to enforce} structural constraints. Considering $w'_i$ as the blending weights from the warped alpha layers $\alpha'$ (from Equation~\ref{eq:blendingweights}), the predicted target image and the regressed depth map (w.r.t. the source) are obtained by:
\begin{equation}
    \hat{\mathbf{I}}_t = \sum_{i=1}^{D}({c'_i}{w'_i}),\quad%
    \hat{\mathbf{D}}_s = \sum_{i=1}^{D}({d^{-1}_i\alpha_i}{w_i}),
    \label{eq:compositing-fw}
\end{equation}
where $d^{-1}_i$ is the disparity of the $i^{\text{th}}$ plane (we use disparity and depth interchangeably).

\subsection{Proposed approach}

In our method, we assume that the ground truth for the target image $\mathbf{I}_t$ is not available. Although this is a common scenario, where stereo data is not available for training, this case cannot be handled by previous approaches based on MPI learning, as discussed in Section~\ref{sec:relatedwork}. For making it possible to train our model from end-to-end and without requiring stereo data, we propose two key strategies, described as follows.

\textit{First}, we warp the source image into the target viewpoint, considering a Lambertian projection of non-occluded regions, based on the corresponding depth map and using morphological filtering to reduce artifacts. This allows us to recover an approximation of $\mathbf{I}_t$, designated by $\mathbf{I}'_t$, which has \rev{been} demonstrated by our experiments to be sufficient and effective for learning an MPI representation that encodes non-occluded regions from the source image. \rev{This step is designated as \textit{forward pass} in Fig.~\ref{fig:proposed-method}}.
\textit{Second}, in order to enforce the network to learn to generate color textures for occluded regions, we propose a training scheme based on cycle consistency: the neural network should be able to predict an MPI from the \textit{target image} that allows recovering the \textit{source image}.
In other words, when considering a cycle consistency training, instead of only feeding the source image as input to the neural network, we also input the warped images $\mathbf{I}'_t$ or $\hat{\mathbf{I}}_t$ (simply referred as $\mathbf{I}^{(\cdot)}_t$) into $f_{\theta}$, which produces an MPI w.r.t. the \textit{target viewpoint}, \rev{which corresponds to the \textit{backward pass} in Fig.~\ref{fig:proposed-method}}. Then, this MPI is projected back to the \textit{source viewpoint}, where the supervision signal is the \textit{source image} $\mathbf{I}_s$.

Note that both the warped image $\mathbf{I}'_t$ and the predicted view $\hat{\mathbf{I}}_t$ can be used as input in this process. To make the difference clear, we refer to the former as \textit{inverse projection training} and to the latter as \textit{cyclic training}.
This strategy enforces the network to learn to inpaint the occluded regions since the source image has all the pixel information that was occluded in $\mathbf{I}^{(\cdot)}_t$. This process is illustrated in Fig.~\ref{fig:proposed-method}, and the details of each part are presented next.

\subsubsection{Self-supervision signal}
\label{sec:self-supervision-signal}

\begin{figure}[!htbp]
\centering
\begin{tabular}{cccccc}
\includegraphics[height=2.0cm]{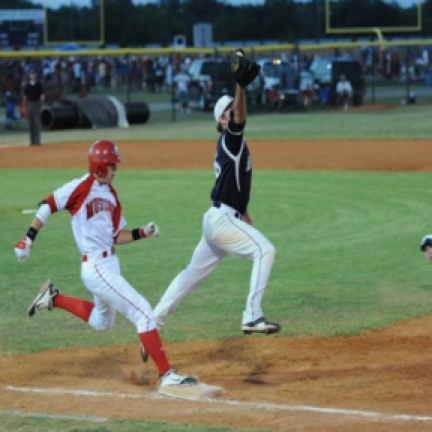}&
\hspace{-0.31cm}\includegraphics[height=2.0cm]{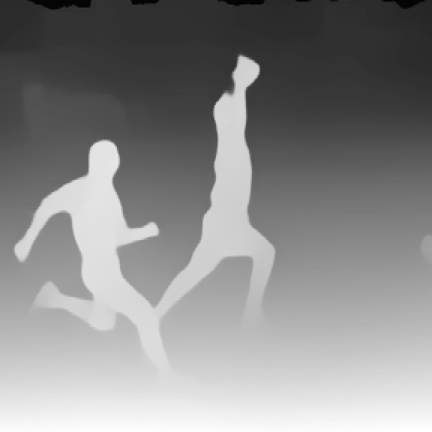}&
\hspace{-0.31cm}\includegraphics[height=2.0cm]{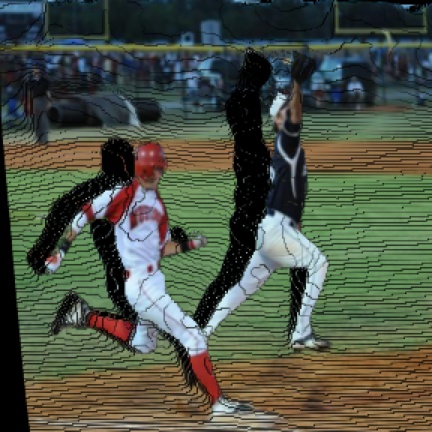}&
\hspace{-0.31cm}\includegraphics[height=2.0cm]{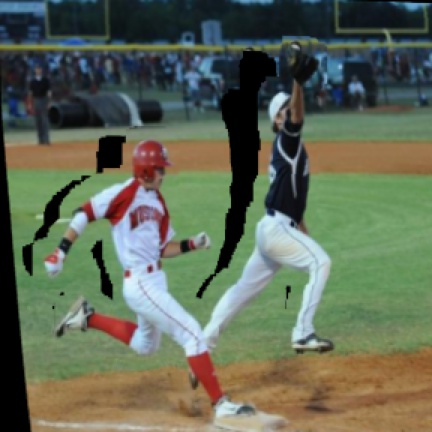}&
\hspace{-0.31cm}\includegraphics[height=2.0cm]{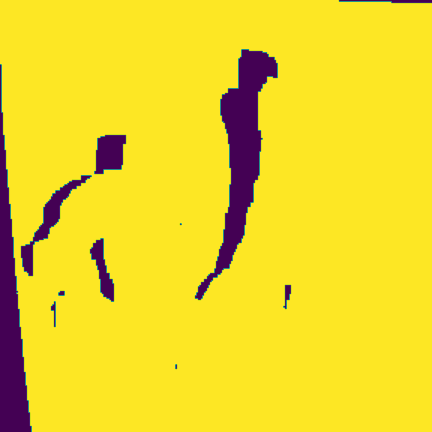}&
\hspace{-0.31cm}\includegraphics[height=2.0cm]{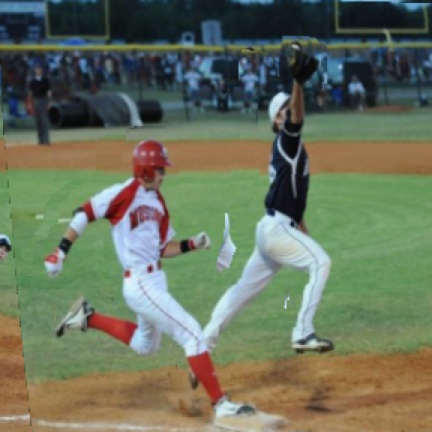}\\
(a)&\hspace{-0.31cm}(b)&\hspace{-0.31cm}(c)&\hspace{-0.31cm}(d)&\hspace{-0.31cm}(e)&\hspace{-0.31cm}(f)
\end{tabular}
\caption{Reconstruction of the target image from the source image (a) and
depth map (b); (c) naive forward projection and (d) forward projection
after morphological filtering; (e) the resulting mask of valid pixels and
(f) the reconstructed approximation $\mathbf{I}'_t$ of the target image.
}
\label{fig:self-supervision-signal}
\end{figure}

Let us consider a source image $\mathbf{I}_s$ captured from viewpoint $v_s$. Our goal in this step is to reconstruct an approximation of the target image $\mathbf{I}_t$ from viewpoint $v_t$, so this reconstructed image could be used as an auxiliary supervision signal. For this, we consider a depth map $\mathbf{D}_s$ as additional information from the source viewpoint, which can be obtained with state-of-the-art monocular depth estimation models (e.g.,~\cite{Ranftl2021}). Then, each pixel from the source image is warped to the target viewpoint $v_t$, as illustrated in Fig.~\ref{fig:self-supervision-signal}.

\noindent
\textbf{Image warping and filtering}.
A naive forward warping would result in sparkle image artifacts or ``flying pixels'', as shown in Fig.~\ref{fig:self-supervision-signal}~(c). Instead, we propose an alternative warping process, where the source image is split into an MPI-like representation, directly obtained by slicing the image accordingly to the depth map. More specifically, we quantize the depth map $\mathbf{D}_s$ into integer values $i \in [1,\dots,D]$ expanding it as a binary representation $\mathbf{q}$ of shape $H\times{}W\times{}D$. Then, each binary mask $\mathbf{q}_i$ (of shape $H\times{}W$) is filtered by a sequence of morphological operations to remove flying pixels and to expand the boundaries of connected regions by a few pixels. In practice, we perform a morphological erosion followed by a dilation, with equivalent filters of size $3\times3$ and $7\times7$, respectively. Note that the filters' size might be adapted to different image resolutions.

\noindent
\textbf{Background filling}.
The warped image and its mask (Fig.~\ref{fig:self-supervision-signal} d-e) would be enough for training a network with L1 loss in the forward pass. However, we may need additional loss functions to improve the quality of predictions, e.g., the VGG perceptual loss~\cite{johnson2016perceptual}. For this, we require \rev{a} dense target and predicted images. We followed the strategy proposed in \cite{watson-2020-stereo-from-mono} for filling in the holes in the warped images with real image samples. In our case, we use the source image of the same sample flipped horizontally. Although this strategy is very simple and efficient, the results are generally plausible in a global perspective, as in Fig.~\ref{fig:self-supervision-signal} (f).

\subsubsection{Training with reverse projection and cyclic loss}

The approximation of the target image reconstructed by our method does not provide a direct supervision signal for learning to inpaint occluded regions in the scene, since it does not have information corresponding to the occluded pixels in the source image. To handle this, we propose a new training strategy, where the input of the network is the approximation $\mathbf{I}'_t$ or the prediction $\hat{\mathbf{I}}_t$, and $\mathbf{I}_s$ is used for supervision. In this case, we can re-write Equation~\ref{eq:mpi-gen} as:
\begin{equation}
    \{(c^t_i, \alpha^t_i)\}_{i=1}^D=f_{\theta}(\mathbf{I}^{(\cdot)}_t),
    \label{eq:mpi-gen-tgt}
\end{equation}
where $c^t_i$ and $\alpha^t_i$ are the color and alpha values w.r.t. the target viewpoint. The same process of warping and compositing an MPI, described in Section~\ref{sec:background-mpi}, can be applied to the new $(c^t_i, \alpha^t_i)$ values, considering a proper swapping between the source and target viewpoints, which results in the MPI $\{(c'^t_i, \alpha'^t_i)\}_{i=1}^D$. Finally, we can render the estimated image in the \textit{source viewpoint} and the depth map in the \textit{target viewpoint} (from Equation~\ref{eq:compositing-fw}):
\begin{equation}
    \hat{\mathbf{I}}_s = \sum_{i=1}^{D}({c'^t_i}{w'^t_i}),\quad%
    \hat{\mathbf{D}}_t = \sum_{i=1}^{D}({d^{-1}_i\alpha^t_i}{w^t_i}),
    \label{eq:compositing-bw}
\end{equation}
where $c'^t_i$ and $w'^t_i$ represent the color values and blending weights after warping from the \textit{target} to the \textit{source} viewpoint, and $w^t_i$ is the blending weights from the target viewpoint.

\noindent
\textbf{Depth map warping and filtering}.
From Equation~\ref{eq:compositing-bw}, we can supervise $\hat{\mathbf{I}}_s$ with the source image $\mathbf{I}_s$. However, a supervision signal is not available for $\hat{\mathbf{D}}_t$. A naive approach would be to estimate $\mathbf{D}_t$ using a monocular depth estimation method from $\mathbf{I}'_t$. This approach would be computationally expensive during training. Instead, we warp the depth map $\mathbf{D}_s$ to the target viewpoint, using a similar warping approach as described in Section~\ref{sec:self-supervision-signal}. In this case, we performed a morphological erosion and dilation with equivalent filters of size $3\times3$ and $5\times5$. This process results in $\mathbf{D}'_t$, as illustrated in Fig.~\ref{fig:proposed-method}.

\subsubsection{Neural network model}
We implement the neural network $f_{\theta}$ with the EfficientNet model~\cite{tan2019efficientnet} cut at block 5 as encoder and a sequence of upsampling and depth-wise convolutions as the decoder. The main outputs of the network are i) the background image with shape $H\times{}W\times3$ and ii) the alpha values with shape $H\times{}W\times(D-1)$, since the last alpha value is always opaque. In addition, the network also outputs intermediate depth predictions at scales $H/2\times{}W/2$ and $H/4\times{}W/4$, which are also supervised during training to help the network to learn the structure of the scene with multi-resolution supervision. The intermediate depth predictions are performed by casting depth regression as a multi-layer classification problem, similar to~\cite{bhat2020adabins}.
More details about the network implementation and the intermediate supervisions are in the supplementary material.

\subsubsection{Training and loss functions}

The network $f_{\theta}$ can be trained in three different ways: i) direct projection, where the source image $\mathbf{I}_s$ is used as input and $\mathbf{I}'_t$ and $\mathbf{D}_s$ are used as supervisions; ii) inverse projection, where the reconstructed target image $\mathbf{I}'_t$ is used as input and $\mathbf{I}_s$ and $\mathbf{D}'_t$ are used as supervision, or iii) cyclic training, where we first perform a direct projection, then we do an inverse projection using as input to the neural network the predicted view $\hat{\mathbf{I}}_t$.
In practice, when we refer to \textit{inverse projection training}, we alternate between direct projection and inverse projection randomly with equal probability. During training, we also generate random target viewpoints $v_t$ (assuming $v_s$ as the identity) with small camera movements.

For supervising the estimated depth maps, we use a common loss function composed of an L1 plus multi-scale gradient matching terms~\cite{MegaDepthLi18}:
\begin{equation}
    \mathcal{L}_{depth}=\|\mathbf{D} - \hat{\mathbf{D}}\|%
    +\alpha\sum_k(\|\nabla_x(\mathbf{D}^k - \hat{\mathbf{D}}^k)\| + \|\nabla_y(\mathbf{D}^k - \hat{\mathbf{D}}^k)\|),
    \label{eq:depth-loss}
\end{equation}
where $\mathbf{D}$ and $\hat{\mathbf{D}}$ are the reference and predicted depths, and $k$ is the scale. Similarly to~\cite{MegaDepthLi18}, we used $k=4$ scales and $\alpha=0.5$.

For supervising the image reconstruction, we used three different losses:
\begin{equation}
    \mathcal{L}_{pix}=\|\mathbf{I} - \hat{\mathbf{I}}\|,\quad%
    \mathcal{L}_{vgg}=\sum_j\|\Phi_j(\mathbf{I}) - \Phi_j(\hat{\mathbf{I}})\|,\quad%
    \mathcal{L}_{style}=\sum_j\|G^{\Phi}_j(\mathbf{I}) - G^{\Phi}_j(\hat{\mathbf{I}})\|,
    \label{eq:image-loss}
\end{equation}
where $\mathbf{I}$ and $\hat{\mathbf{I}}$ are the reference and predicted views, $\Phi$ is the 16-layer VGG model, $G^{\Phi}$ is the Gram matrix from VGG features, and $j$ indexes the first three VGG blocks. Please refer to~\cite{johnson2016perceptual} for more details about VGG and style losses. The final loss is given by:
\begin{equation}
    \mathcal{L}_{total}=\mathcal{L}_{depth}+\mathcal{L}_{pix}+\beta\mathcal{L}_{vgg}+\gamma\mathcal{L}_{style}
    \label{eq:total-loss}
\end{equation}

\section{Experiments}
\label{sec:experiments}

In this section, we evaluated our method considering a challenging case where no stereo data is available for training, i.e., our model is trained using a single view dataset with no ground truth depth nor target images, and is evaluated on well-known stereo datasets from the literature, which characterizes a zero-shot evaluation scenario.

\subsection{Datasets and evaluation metrics}

\noindent
\textbf{Training datasets}.
For training, we consider two different datasets: Places II~\cite{zhou2017places} and a dataset with images from Flickr~\cite{luvizon2021adaptive}, here referred to as Flickr-Pictures dataset. The former is composed of about 1.8M images and the latter has about 95K images in high resolution, which were resized to $1024\times768$ at least, keeping the original ratio. For both datasets, we computed the depth maps for supervision using the available model from~\cite{Ranftl2021}.

\noindent
\textbf{Evaluation datasets}.
RealEstate10K~\cite{zhou2018stereo} is a common dataset for novel view synthesis, composed of 70K sequences extracted from YouTube videos. We used the evaluation set composed of 1500 testing triplets proposed by~\cite{Shih3DP20}, choosing the source and target images to be 6 and 10 frames apart. We also evaluated our method on the Mannequin Challenge dataset~\cite{Zhengqi_CVPR_2019}, which is traditionally a depth estimation dataset based on structure-from-motion (SfM). This is a challenging dataset since it contains complex scenes with many people and often with blurry images. \rev{The} evaluation was performed on the official test split, choosing the source and target frames to be 3 frames apart.


\subsection{Implementation details}
\label{sec:imp-details}

We trained our model using Adam as optimizer, with parameters $\beta_1=0.9$,  $\beta_2=0.999$, and learning rate $10^{-3}$, with batches of 4 images for about 10 million iterations. We set the optimization parameters of our method as $\beta=0.01$ and $\gamma=0.0001$, \rev{according} to our ablation studies in the Appendices.

\begin{table}[]
    \centering
    \begin{tabular}{c c c c c c c}
        \toprule
        & \multicolumn{2}{c}{SSIM $\uparrow$} & \multicolumn{2}{c}{PSNR $\uparrow$} & \multicolumn{2}{c}{LPIPS $\downarrow$}  \\
        \cmidrule(lr){2-3}
        \cmidrule(lr){4-5}
        \cmidrule(lr){6-7}
        Method & $n=6$ & $n=10$ & $n=6$ & $n=10$ & $n=6$ & $n=10$ \\
        \midrule
        Single-View Synthesis~\cite{Tucker_2020_CVPR}  & 0.844 & 0.795 & 23.50 & 20.80 & 0.138 & 0.209 \\
        3D-Photography~\cite{Shih3DP20} & {0.859} & {0.795} & {24.30} & {21.27} & {0.087} & {0.141} \\
        Adaptive-MPI~\cite{luvizon2021adaptive} & {0.804} & {0.735} & {21.09} & {18.23} & {0.113} & {0.187} \\
        \midrule
        Ours & {0.842} & {0.786} & {22.79} & {19.96} & {0.115} & {0.176} \\ 
        \rev{Ours + fine-tuning on RE10K} & \rev{0.845} & \rev{0.788} & \rev{22.85} & \rev{19.89} & \rev{0.129} & \rev{0.196} \\ 
        \bottomrule
    \end{tabular}
    \caption{Comparison with previous methods on RealEstate10K. \rev{We fine-tuned our model using 10\% of training images from RE10K (without ground-truth depths)}. Note that \cite{Tucker_2020_CVPR} was trained on RE10K \rev{with ground-truth depths} and \cite{Shih3DP20} uses multiple views during inference to estimate depth.}
    \label{tab:real_estate}
\end{table}

\begin{table}[]
    \centering
    \begin{tabular}{c c c c c}
        \toprule
        Method & FID $\downarrow$ & SSIM $\uparrow$ & PSNR $\uparrow$ & LPIPS $\downarrow$ \\
        \midrule
        Single-View Synthesis~\cite{Tucker_2020_CVPR} & 21.24 & 0.617 & 13.56 & 0.514 \\
        Ours  & 16.10 & 0.633 & 14.02 & 0.479 \\ 
        \bottomrule
    \end{tabular}
    \caption{Comparison of MPI end-to-end approaches on Mannequin Challenge dataset.}
    \label{tab:mannequin}
\end{table}

\subsection{Comparison with previous methods}

\noindent
\textbf{RealEstate10K}.
We compare our method with~\cite{Tucker_2020_CVPR}, \cite{Shih3DP20}, and \cite{luvizon2021adaptive} on RealEstate10K, evaluating the SSIM~\cite{Zhou_SSIM_2004}, PSNR, and LPIPS~\cite{zhang2018perceptual} metrics between the images produced by each method and the ground truth. The results are presented in Table~\ref{tab:real_estate}. We replicated the results from all previous methods using the author's implementation and pre-trained models. Our results are close to the state of the art, even considering that \cite{Tucker_2020_CVPR} was trained on RealEstate10K and \cite{Shih3DP20} uses multiple views for depth estimation, while our method was trained only with single images from Places II and does not \rev{use} multiple views during inference. \rev{We also fine-tuned our model using 10\% of training images from RealEstate10K, which resulted in a \rev{slight} improvement in the SSIM metric}.

\noindent
\textbf{Mannequin Challenge}.
One advantage of our self-supervised method is its capability to generalize on challenging scenarios. To demonstrate this, we evaluated our method on the Mannequin Challenge, which contains complex scenes with people and moving objects, and is more than fifty times smaller than RealEstate10K, making it unfeasible for training deep networks for view synthesis. Another challenge in this dataset is the wide variety of scene scales, making it hard for any scale-free 3D representation to generate novel views correctly aligned with the ground truth data. We refer to the Appendix for examples of the misalignment problem. For this reason, we also reported results using the Fréchet Inception Distance (FID)~\cite{heusel2017gans}, an evaluation metric suited for ill-posed problems, to deal with misalignment issues. We focus on MPI models trained end-to-end, comparing our self-supervised model trained on Places II with Single-View Synthesis~\cite{Tucker_2020_CVPR} trained on RealEstate10K. Results on Table \ref{tab:mannequin} show the superiority of our model in this challenging case.

\begin{table}[]
\centering
\small
\begin{tabular}{@{}cccccc|ccc@{}}
\toprule
\multicolumn{6}{c|}{Training strategy \rev{on Places II}} & \multicolumn{3}{c}{Validation on RE10K} \\ \midrule
$\mathcal{L}_{depth}$      & $\mathcal{L}_{pix}$ & $\mathcal{L}_{vgg}$ & $\mathcal{L}_{style}$ & Inverse proj.     & Cyclic & SSIM $\uparrow$ & PSNR $\uparrow$ & LPIPS $\downarrow$ \\
\midrule
\checkmark &            &            &            &   &        & 0.734 & 17.699 & 0.237 \\ 
           & \checkmark &        &            &       &    & 0.758 & 19.349 & 0.280 \\ 
\checkmark & \checkmark &        &            &         &  & 0.760 & 19.473 & 0.215 \\ 
\checkmark & \checkmark & \checkmark & &     & & 0.752 & 19.332 & 0.195 \\ 
\checkmark & \checkmark & \checkmark & \checkmark  &  & & 0.735 & 18.748 & 0.183 \\ 
\checkmark & \checkmark & \checkmark & \checkmark & \checkmark & & 0.761 & 19.556 & \textbf{0.182} \\ 
\checkmark & \checkmark & \checkmark & \checkmark  &  & \checkmark & \textbf{0.765} & \textbf{19.773} & \textbf{0.182} \\ 
\bottomrule
\end{tabular}
\caption{Ablation study considering different training strategies in our method.}
\label{tab:ablation}
\end{table}

\subsection{Ablation studies}

We also performed ablation experiments with different training strategies of our method to better understand their relative importance, which is presented in Table~\ref{tab:ablation}. \rev{In all these experiments, we trained our model with images from Places II for 500K iterations, using a fixed image resolution of $384\times288$ pixels.} We tested several combinations of the loss functions in the final loss, considering inverse projection and cyclic training. We can notice that depth and pixel losses are complementary for learning. Although these two losses alone already resulted in satisfying SSIM and PSNR metrics, the LPIPS result is poor, which reflects the low texture quality of the generated views. For this, we included the VGG and style losses, which increase LPIPS metric substantially. \rev{From our experiments, we noticed that LPIPS is very sensitive to blurry images, therefore this metric is relevant for more realistic image predictions.}
We can observe that both inverse projection and cyclic training have improved all the metrics consistently, which supports our main idea.

\rev{In order to better understand the proposed cyclic training strategy in regard of the availability of target views and different depth estimation models, we performed additional experiments in Table~\ref{tab:ablation2}. First, we trained our model on RE10K considering a simple forward training (first row) and cyclic training, which improved all the metrics. One could also argue that the DPT model~\cite{Ranftl2021} was trained with multiple datasets, including stereo data. We show that our approach does not depend on stereo data by comparing DPT and AdaBins~\cite{bhat2020adabins},  which was trained NYU depth, a Kinect based single depth dataset. From this evaluation, we can observe that both models resulted in close SSIM and PSNR metrics, although AdaBins degraded the LPIPS metric.}

\begin{figure}[!htbp]
\centering
\includegraphics[height=3.4cm]{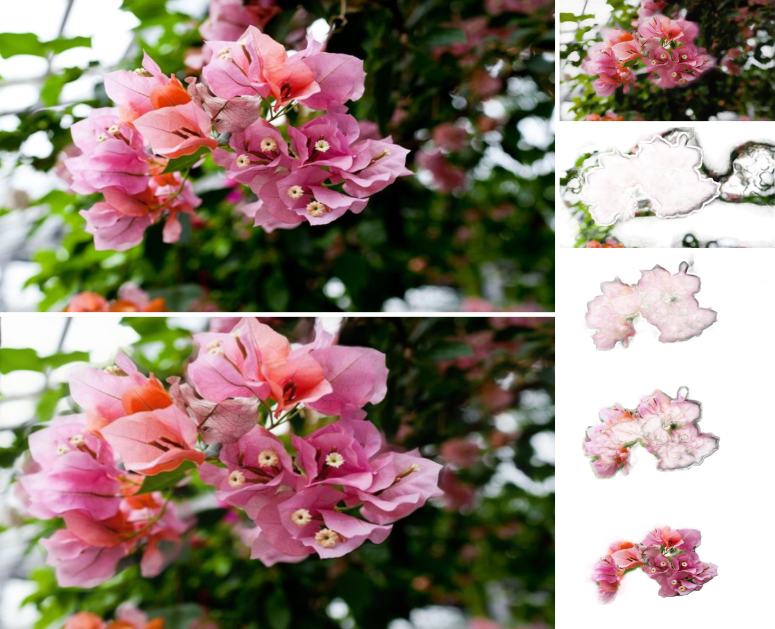}
\includegraphics[height=3.4cm]{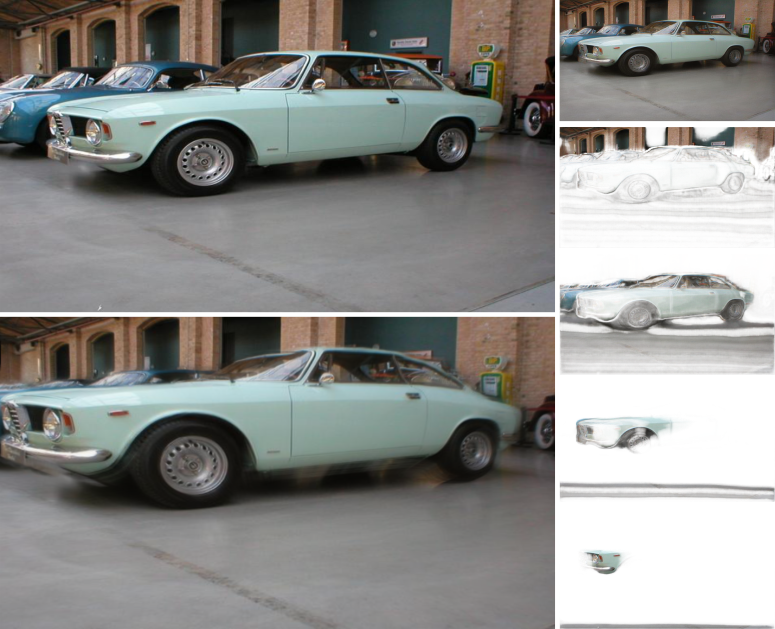}
\includegraphics[height=3.4cm]{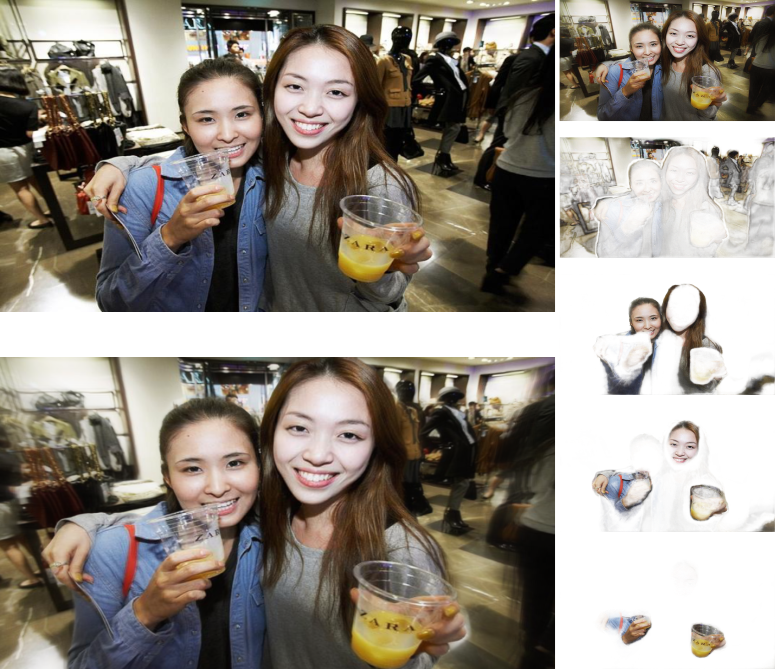}
\caption{Results on Places II: source image (top), new view (bottom), and selected planes.
}
\label{fig:qualitative-results}
\end{figure}

Finally, we presented some qualitative results from our method in Fig.~\ref{fig:qualitative-results}. This evidences the capability of our approach to generalize well on different and challenging types of images.
\rev{The main failures of our method are due to wrong depth predictions of our model, which is a consequence of incorrect pseudo ground-truth depth used during training. Blurry images are also observed for large baselines, which is a common problem on MPI based methods. Please see the supplementary material and video demonstration of the method for more results.}


\begin{table}[]
\centering
\small
\begin{tabular}{@{}ccc|ccc@{}}
\toprule
\multicolumn{3}{c|}{Training strategy} & \multicolumn{3}{c}{Validation on RE10K} \\ \midrule
Training dataset & Depth model & Cyclic & SSIM $\uparrow$ & PSNR $\uparrow$ & LPIPS $\downarrow$ \\
\midrule
RE10K & DPT &            & 0.767 & 20.729 & 0.207 \\ 
RE10K & DPT & \checkmark & 0.772 & 20.959 & 0.181 \\ 
\midrule
Places365 & DPT &            & 0.735 & 18.748 & 0.183 \\ 
Places365 & DPT & \checkmark & 0.765 & 19.773 & 0.182 \\ 
Places365 & AdaBins & \checkmark & 0.760 & 19.948 & 0.229 \\ 
\bottomrule
\end{tabular}
\caption{Comparison of different training data and depth maps by DPT~\cite{Ranftl2021} and AdaBins~\cite{bhat2020adabins}.}
\label{tab:ablation2}
\end{table}




\section{Conclusions}
\label{sec:conclusion}

In this paper, we presented a new method for learning an MPI representation with \rev{intermediate depth and cyclic supervision}. The proposed method can be trained with single view data by exploring inverse projection and cyclic training strategies for learning to inpaint occluded regions in the scene. We compared our approach with recent methods for single view synthesis from the literature. Our method was evaluated in a zero-shot scenario and achieved results comparable to the results in the state of the art, even when compared to methods trained on the target datasets. \rev{We also presented ablation studies to show the relative importance of each component of the proposed framework, which evidenced the benefits of our method}.

\bibliography{references}

\begin{thebibliography}{35}
\providecommand{\natexlab}[1]{#1}
\providecommand{\url}[1]{\texttt{#1}}
\expandafter\ifx\csname urlstyle\endcsname\relax
  \providecommand{\doi}[1]{doi: #1}\else
  \providecommand{\doi}{doi: \begingroup \urlstyle{rm}\Url}\fi

\bibitem[Aleotti et~al.(2021)Aleotti, Poggi, and
  Mattoccia]{aleotti2021learning}
Filippo Aleotti, Matteo Poggi, and Stefano Mattoccia.
\newblock Learning optical flow from still images.
\newblock In \emph{Proceedings of the IEEE/CVF Conference on Computer Vision
  and Pattern Recognition}, pages 15201--15211, 2021.

\bibitem[Alhashim and Wonka(2018)]{Alhashim2018}
Ibraheem Alhashim and Peter Wonka.
\newblock High quality monocular depth estimation via transfer learning.
\newblock \emph{arXiv e-prints}, abs/1812.11941:\penalty0 arXiv:1812.11941,
  2018.

\bibitem[Bhat et~al.(2021)Bhat, Alhashim, and Wonka]{bhat2020adabins}
Shariq~Farooq Bhat, Ibraheem Alhashim, and Peter Wonka.
\newblock Adabins: Depth estimation using adaptive bins.
\newblock In \emph{Proceedings of the IEEE/CVF Conference on Computer Vision
  and Pattern Recognition (CVPR)}, pages 4009--4018, June 2021.

\bibitem[Chen et~al.(2016)Chen, Fu, Yang, and Deng]{chen2016depthwild}
Weifeng Chen, Zhao Fu, Dawei Yang, and Jia Deng.
\newblock Single-image depth perception in the wild.
\newblock In D.~Lee, M.~Sugiyama, U.~Luxburg, I.~Guyon, and R.~Garnett,
  editors, \emph{Advances in Neural Information Processing Systems}, volume~29.
  Curran Associates, Inc., 2016.

\bibitem[Choi et~al.(2018)Choi, Choi, Kim, Ha, Kim, and Choo]{choi2018stargan}
Yunjey Choi, Minje Choi, Munyoung Kim, Jung-Woo Ha, Sunghun Kim, and Jaegul
  Choo.
\newblock Stargan: Unified generative adversarial networks for multi-domain
  image-to-image translation.
\newblock In \emph{Proceedings of the IEEE Conference on Computer Vision and
  Pattern Recognition}, 2018.

\bibitem[Dosovitskiy et~al.(2020)Dosovitskiy, Beyer, Kolesnikov, Weissenborn,
  Zhai, Unterthiner, Dehghani, Minderer, Heigold, Gelly,
  et~al.]{dosovitskiy2020image}
Alexey Dosovitskiy, Lucas Beyer, Alexander Kolesnikov, Dirk Weissenborn,
  Xiaohua Zhai, Thomas Unterthiner, Mostafa Dehghani, Matthias Minderer, Georg
  Heigold, Sylvain Gelly, et~al.
\newblock An image is worth 16x16 words: Transformers for image recognition at
  scale.
\newblock \emph{arXiv preprint arXiv:2010.11929}, 2020.

\bibitem[Eigen et~al.(2014)Eigen, Puhrsch, and Fergus]{eigen2014depth}
David Eigen, Christian Puhrsch, and Rob Fergus.
\newblock Depth map prediction from a single image using a multi-scale deep
  network.
\newblock In \emph{Advances in Neural Information Processing Systems}, pages
  2366--2374, 2014.

\bibitem[Hani et~al.(2020)Hani, Engin, Chao, and Isler]{hani2020object}
Nicolai Hani, Selim Engin, Jun-Jee Chao, and Volkan Isler.
\newblock Continuous object representation networks: Novel view synthesis
  without target view supervision.
\newblock In H.~Larochelle, M.~Ranzato, R.~Hadsell, M.~F. Balcan, and H.~Lin,
  editors, \emph{Advances in Neural Information Processing Systems}, volume~33,
  pages 6086--6099. Curran Associates, Inc., 2020.

\bibitem[Hartley and Zisserman(2004)]{Hartley2004}
R.~I. Hartley and A.~Zisserman.
\newblock \emph{Multiple View Geometry in Computer Vision}.
\newblock Cambridge University Press, ISBN: 0521540518, second edition, 2004.

\bibitem[Heusel et~al.(2017)Heusel, Ramsauer, Unterthiner, Nessler, and
  Hochreiter]{heusel2017gans}
Martin Heusel, Hubert Ramsauer, Thomas Unterthiner, Bernhard Nessler, and Sepp
  Hochreiter.
\newblock Gans trained by a two time-scale update rule converge to a local nash
  equilibrium.
\newblock \emph{arXiv preprint arXiv:1706.08500}, 2017.

\bibitem[Huang et~al.(2020)Huang, Tseng, Lee, and Huang]{Huang_ECCV2020}
Hsin-Ping Huang, Hung-Yu Tseng, Hsin-Ying Lee, and Jia-Bin Huang.
\newblock Semantic view synthesis.
\newblock In Andrea Vedaldi, Horst Bischof, Thomas Brox, and Jan-Michael Frahm,
  editors, \emph{Computer Vision -- ECCV 2020}, pages 592--608, Cham, 2020.
  Springer International Publishing.
\newblock ISBN 978-3-030-58610-2.

\bibitem[Johnson et~al.(2016)Johnson, Alahi, and
  Fei-Fei]{johnson2016perceptual}
Justin Johnson, Alexandre Alahi, and Li~Fei-Fei.
\newblock Perceptual losses for real-time style transfer and super-resolution.
\newblock In \emph{European conference on computer vision}, pages 694--711.
  Springer, 2016.

\bibitem[Kim et~al.(2017)Kim, Cha, Kim, Lee, and Kim]{kim2017learning}
Taeksoo Kim, Moonsu Cha, Hyunsoo Kim, Jung~Kwon Lee, and Jiwon Kim.
\newblock Learning to discover cross-domain relations with generative
  adversarial networks, 2017.

\bibitem[Li and Snavely(2018)]{MegaDepthLi18}
Zhengqi Li and Noah Snavely.
\newblock Megadepth: Learning single-view depth prediction from internet
  photos.
\newblock In \emph{Computer Vision and Pattern Recognition (CVPR)}, 2018.

\bibitem[Li et~al.(2019)Li, Dekel, Cole, Tucker, Snavely, Liu, and
  Freeman]{Zhengqi_CVPR_2019}
Zhengqi Li, Tali Dekel, Forrester Cole, Richard Tucker, Noah Snavely, Ce~Liu,
  and William~T. Freeman.
\newblock Learning the depths of moving people by watching frozen people.
\newblock In \emph{Proceedings of the IEEE Conference on Computer Vision and
  Pattern Recognition (CVPR)}, 2019.

\bibitem[Luvizon et~al.(2021)Luvizon, Carvalho, dos Santos, Conceicao,
  Flores-Campana, Decker, Souza, Pedrini, Joia, and
  Penatti]{luvizon2021adaptive}
Diogo~C Luvizon, Gustavo Sutter~P Carvalho, Andreza~A dos Santos, Jhonatas~S
  Conceicao, Jose~L Flores-Campana, Luis~GL Decker, Marcos~R Souza, Helio
  Pedrini, Antonio Joia, and Otavio~AB Penatti.
\newblock Adaptive multiplane image generation from a single internet picture.
\newblock In \emph{Proceedings of the IEEE/CVF Winter Conference on
  Applications of Computer Vision}, pages 2556--2565, 2021.

\bibitem[Mildenhall et~al.(2020)Mildenhall, Srinivasan, Tancik, Barron,
  Ramamoorthi, and Ng]{mildenhall2020nerf}
Ben Mildenhall, Pratul~P. Srinivasan, Matthew Tancik, Jonathan~T. Barron, Ravi
  Ramamoorthi, and Ren Ng.
\newblock Nerf: Representing scenes as neural radiance fields for view
  synthesis.
\newblock In \emph{ECCV}, 2020.

\bibitem[Niklaus et~al.(2019)Niklaus, Mai, Yang, and Liu]{Niklaus_TOG_2019}
Simon Niklaus, Long Mai, Jimei Yang, and Feng Liu.
\newblock 3d ken burns effect from a single image.
\newblock \emph{ACM Transactions on Graphics}, 38\penalty0 (6):\penalty0
  184:1--184:15, 2019.

\bibitem[Porter and Duff(1984)]{Porter_OverComposite_ACM_1984}
Thomas Porter and Tom Duff.
\newblock Compositing digital images.
\newblock In \emph{Proceedings of the 11th Annual Conference on Computer
  Graphics and Interactive Techniques}, page 253–259. ACM, 1984.

\bibitem[Ranftl et~al.(2020)Ranftl, Lasinger, Hafner, Schindler, and
  Koltun]{Ranftl2020}
Ren\'{e} Ranftl, Katrin Lasinger, David Hafner, Konrad Schindler, and Vladlen
  Koltun.
\newblock Towards robust monocular depth estimation: Mixing datasets for
  zero-shot cross-dataset transfer.
\newblock \emph{IEEE Transactions on Pattern Analysis and Machine Intelligence
  (TPAMI)}, 2020.

\bibitem[Ranftl et~al.(2021)Ranftl, Bochkovskiy, and Koltun]{Ranftl2021}
Ren\'{e} Ranftl, Alexey Bochkovskiy, and Vladlen Koltun.
\newblock Vision transformers for dense prediction.
\newblock \emph{ArXiv preprint}, 2021.

\bibitem[Schonberger and Frahm()]{schonberger_cvpr_2016}
Johannes~L. Schonberger and Jan-Michael Frahm.
\newblock Structure-from-motion revisited.
\newblock In \emph{2016 {IEEE} Conference on Computer Vision and Pattern
  Recognition ({CVPR})}, pages 4104--4113. {IEEE}.
\newblock ISBN 978-1-4673-8851-1.
\newblock \doi{10.1109/CVPR.2016.445}.

\bibitem[Shade et~al.(1998)Shade, Gortler, He, and Szeliski]{Shade_LDI_1998}
Jonathan Shade, Steven Gortler, Li-wei He, and Richard Szeliski.
\newblock Layered depth images.
\newblock In \emph{Proceedings of the 25th Annual Conference on Computer
  Graphics and Interactive Techniques}, SIGGRAPH '98, page 231–242, New York,
  NY, USA, 1998. Association for Computing Machinery.
\newblock ISBN 0897919998.
\newblock \doi{10.1145/280814.280882}.

\bibitem[Shih et~al.(2020)Shih, Su, Kopf, and Huang]{Shih3DP20}
Meng-Li Shih, Shih-Yang Su, Johannes Kopf, and Jia-Bin Huang.
\newblock 3d photography using context-aware layered depth inpainting.
\newblock In \emph{IEEE Conference on Computer Vision and Pattern Recognition
  (CVPR)}, 2020.

\bibitem[Srinivasan et~al.(2019)Srinivasan, Tucker, Barron, Ramamoorthi, Ng,
  and Snavely]{Srinivasan_CVPR_2019}
Pratul~P. Srinivasan, Richard Tucker, Jonathan~T. Barron, Ravi Ramamoorthi, Ren
  Ng, and Noah Snavely.
\newblock Pushing the boundaries of view extrapolation with multiplane images.
\newblock In \emph{2019 IEEE/CVF Conference on Computer Vision and Pattern
  Recognition (CVPR)}, pages 175--184, 2019.
\newblock \doi{10.1109/CVPR.2019.00026}.

\bibitem[Szeliski and Golland(1999)]{szeliski1999stereo}
Richard Szeliski and Polina Golland.
\newblock Stereo matching with transparency and matting.
\newblock \emph{International Journal of Computer Vision}, 32\penalty0
  (1):\penalty0 45--61, 1999.

\bibitem[Tan and Le(2019)]{tan2019efficientnet}
Mingxing Tan and Quoc Le.
\newblock Efficientnet: Rethinking model scaling for convolutional neural
  networks.
\newblock In \emph{International Conference on Machine Learning}, pages
  6105--6114. PMLR, 2019.

\bibitem[Tucker and Snavely(2020)]{Tucker_2020_CVPR}
Richard Tucker and Noah Snavely.
\newblock Single-view view synthesis with multiplane images.
\newblock In \emph{Proceedings of the IEEE/CVF Conference on Computer Vision
  and Pattern Recognition (CVPR)}, June 2020.

\bibitem[Tulsiani et~al.(2018)Tulsiani, Tucker, and Snavely]{lsiTulsiani18}
Shubham Tulsiani, Richard Tucker, and Noah Snavely.
\newblock Layer-structured 3d scene inference via view synthesis.
\newblock In \emph{ECCV}, 2018.

\bibitem[Wang et~al.(2004)Wang, Bovik, Sheikh, and Simoncelli]{Zhou_SSIM_2004}
Zhou Wang, A.C. Bovik, H.R. Sheikh, and E.P. Simoncelli.
\newblock Image quality assessment: from error visibility to structural
  similarity.
\newblock \emph{IEEE Transactions on Image Processing}, 13\penalty0
  (4):\penalty0 600--612, 2004.
\newblock \doi{10.1109/TIP.2003.819861}.

\bibitem[Watson et~al.(2020)Watson, Aodha, Turmukhambetov, Brostow, and
  Firman]{watson-2020-stereo-from-mono}
Jamie Watson, Oisin~Mac Aodha, Daniyar Turmukhambetov, Gabriel~J. Brostow, and
  Michael Firman.
\newblock Learning stereo from single images.
\newblock In \emph{European Conference on Computer Vision ({ECCV})}, 2020.

\bibitem[Zhang et~al.(2018)Zhang, Isola, Efros, Shechtman, and
  Wang]{zhang2018perceptual}
Richard Zhang, Phillip Isola, Alexei~A Efros, Eli Shechtman, and Oliver Wang.
\newblock The unreasonable effectiveness of deep features as a perceptual
  metric.
\newblock In \emph{CVPR}, 2018.

\bibitem[Zhou et~al.(2017)Zhou, Lapedriza, Khosla, Oliva, and
  Torralba]{zhou2017places}
Bolei Zhou, Agata Lapedriza, Aditya Khosla, Aude Oliva, and Antonio Torralba.
\newblock Places: A 10 million image database for scene recognition.
\newblock \emph{IEEE Transactions on Pattern Analysis and Machine
  Intelligence}, 2017.

\bibitem[Zhou et~al.(2018)Zhou, Tucker, Flynn, Fyffe, and
  Snavely]{zhou2018stereo}
Tinghui Zhou, Richard Tucker, John Flynn, Graham Fyffe, and Noah Snavely.
\newblock Stereo magnification: Learning view synthesis using multiplane
  images.
\newblock In \emph{SIGGRAPH}, 2018.

\bibitem[Zhu et~al.(2017)Zhu, Park, Isola, and Efros]{zhu2017unpaired}
Jun-Yan Zhu, Taesung Park, Phillip Isola, and Alexei~A Efros.
\newblock Unpaired image-to-image translation using cycle-consistent
  adversarial networks.
\newblock In \emph{Proceedings of the IEEE international conference on computer
  vision}, pages 2223--2232, 2017.

\end{thebibliography}
\end{document}